\documentclass[10pt,twocolumn,letterpaper]{article}
\usepackage[pagenumbers]{cvpr} 
\usepackage{graphicx}
\usepackage{svg}
\usepackage{amsmath}
\usepackage{amssymb}
\usepackage{booktabs}
\usepackage[ruled,linesnumbered]{algorithm2e}
\usepackage{algorithm2e}
\usepackage{times}
\usepackage{epsfig}
\usepackage{booktabs}
\usepackage{multirow}
\usepackage{makecell}
\usepackage{enumitem}
\usepackage{colortbl}
\usepackage{tabulary}
\usepackage{color} 
\usepackage{bbding}
\usepackage{pifont}
\usepackage{diagbox}
\usepackage{xcolor} 
 \usepackage[square, comma, sort&compress, numbers]{natbib}
\newcommand{\rongyu}[1]{\textcolor{magenta}{#1}}

\usepackage[capitalize]{cleveref}
\crefname{section}{Sec.}{Secs.}
\Crefname{section}{Section}{Sections}
\Crefname{table}{Table}{Tables}
\crefname{table}{Tab.}{Tabs.}

\def\cvprPaperID{257} 
\def\confName{CVPR}
\def\confYear{2023}

\makeatletter
\newbox\abstract@box
\renewenvironment{abstract}
  {\global\setbox\abstract@box=\vbox\bgroup
     \hsize=\textwidth\linewidth=\textwidth
    \small
    \begin{center}%
    {\bfseries \abstractname\vspace{-.5em}\vspace{\z@}}%
    \end{center}%
    \quotation}
  {\endquotation\egroup}
\expandafter\def\expandafter\@maketitle\expandafter{\@maketitle
  \ifvoid\abstract@box\else\unvbox\abstract@box\if@twocolumn\vskip1.5em\fi\fi}
\makeatother

\begin{document}
\title{VeCAF: Vision-language Collaborative Active Finetuning with \\
Training Objective Awareness}
\author{Rongyu Zhang$^{1,2,\ast}$, Zefan Cai$^{2,6,\ast }$, Huanrui Yang$^{3,\ast}$, Zidong Liu$^{4}$, Denis Gudovskiy$^{5}$, Tomoyuki Okuno$^{5}$, \\Yohei Nakata$^{5}$, Kurt Keutzer$^{3}$, Baobao Chang$^{6}$, Yuan Du$^{1,\dag}$, Li Du$^{1}$, Shanghang Zhang$^{2,\dag}$\\
$^{1}$ Nanjing University,
$^{2}$ National Key Laboratory for Multimedia Information Processing, \\School of Computer Science, Peking University,
$^{3}$ University of California, Berkeley,\\
$^{4}$ Tsinghua University,
$^{5}$ Panasonic,
$^{6}$School of Software and Microelectronics, Peking University \\
}
\begin{abstract}
Finetuning a pretrained vision model (PVM) is a common technique for learning downstream vision tasks.
However, the conventional finetuning process with randomly sampled data points results in diminished training efficiency.
To address this drawback, we propose a novel approach, \textbf{V}ision-\textit{languag}\textbf{e} \textbf{C}\textit{ollaborative} \textbf{A}\textit{ctive} \textbf{F}\textit{inetuning} (\texttt{VeCAF}). 
With the emerging availability of labels and natural language annotations of images through web-scale crawling or controlled generation, VeCAF makes use of these information to perform parametric data selection for PVM finetuning. VeCAF incorporates the finetuning objective to select significant data points that effectively guide the PVM towards faster convergence to meet the performance goal. This process is assisted by the inherent semantic richness of the text embedding space which we use to augment image features. 
Furthermore, the flexibility of text-domain augmentation allows VeCAF to handle out-of-distribution scenarios without external data.
Extensive experiments show the leading performance and high computational efficiency of VeCAF that is superior to baselines in both in-distribution and out-of-distribution image classification tasks. 
On ImageNet, VeCAF uses up to 3.3$\times$ less training batches to reach the target performance compared to full finetuning, and achieves an accuracy improvement of 2.7\% over the state-of-the-art active finetuning method with the same number of batches. 
\end{abstract}
\maketitle
\footnotetext{$\ast$ Equal contributions; $\dag$ Corresponding authors}
\section{Introduction}
\begin{figure}[t]
\includegraphics[width=0.95\linewidth]{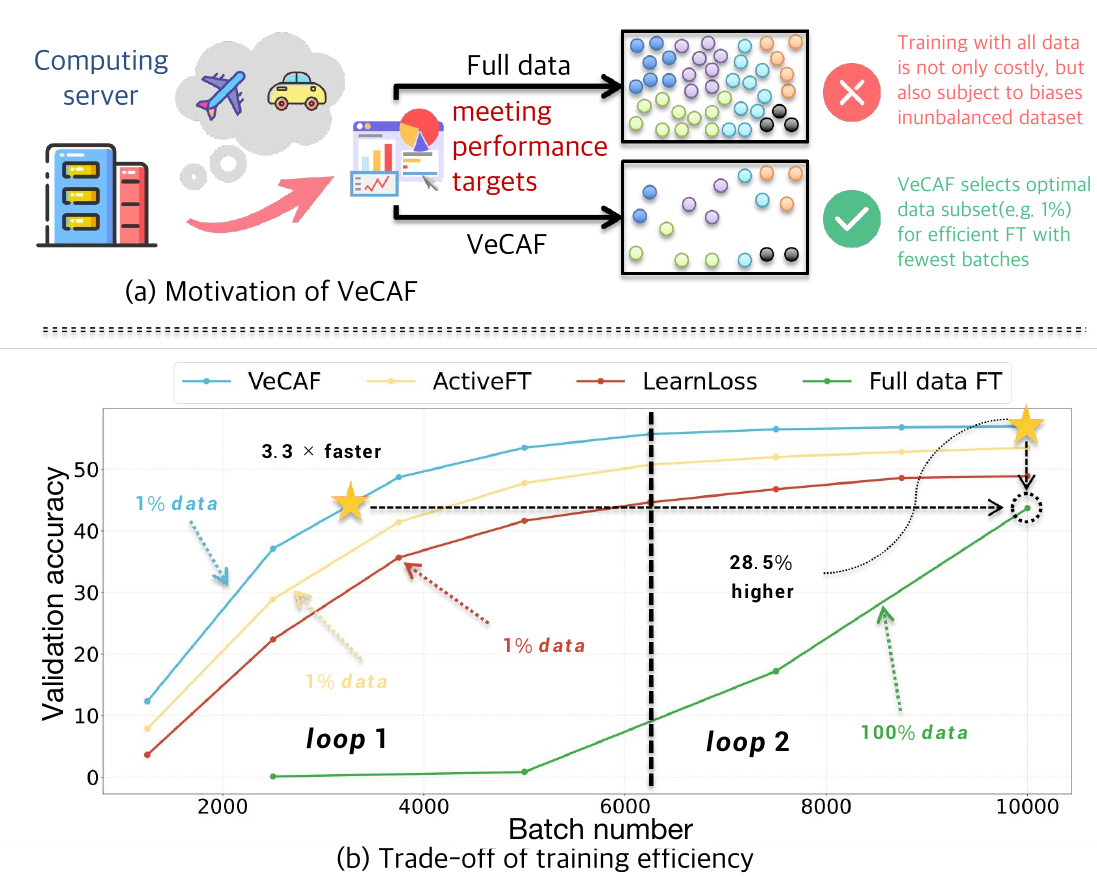}
\centering
\caption{(a) Motivation of VeCAF. We select the optimal subset from a large labeled training set for efficient finetuning (FT) towards a user-specified objective. (b) Training curve comparison on ImageNet-1K validation set. All baselines select 1\% of data in each FT loop with the exception of a conventional setup with full-data FT. VeCAF achieves the target accuracy faster with significantly fewer training batches and achieves higher accuracy with the same training cost.}
\label{fig:teaser}
\end{figure}
Deep learning has made significant progress in the field of computer vision that is typically attributed to the use of large-scale models and datasets~\cite{deng2009imagenet,lin2014microsoft}. Hence, training such models from scratch is a time-consuming process and demands extensive amount of data. To address this, the pretraining-finetuning~\cite{devlin2018bert,touvron2023llama,dosovitskiy2020image,radford2021learning} paradigm has been recognized as a favorable approach for both vision and language tasks. For vision tasks, a model can be first trained on abundant supervised or unsupervised data and be saved as a pretrained vision model (PVM)~\cite{ali2021xcit,liu2021swin,caron2021emerging}. Then, the PVM is finetuned on a labeled dataset for a specific downstream task. By capitalizing on ample pretraining data and conserving valuable training resources during the finetuning stage, this paradigm has archived remarkable adoption in practical applications.



In the real-world deployments, practitioners aim to adapt deep learning models to a certain scenario or tune a model towards a specific performance target with minimal efforts for data selection and with quick training. Training with all available downstream task data can be not only costly but also can lead to a biased or degraded performance in case of improperly collected data.
This motivates the proposal of a data selection framework that can actively select the optimal data subset for finetuning. 
Previous work on active learning~\cite{bengar2021reducing,xie2023active} has shown the feasibility of PVM finetuning with only a small subset (e.g., less than 5\%) of training data while achieving high performance metrics in downstream tasks. 
However, this line of work is often limited by the setting of low label availability which hinders its effectiveness to meet the user-specified objectives.

With the growing feasibility to gather large amounts of images with labels and natural language captions in the target domain through web-scale data crawling~\cite{schuhmann2022laion} or controlled generation~\cite{li2023open}, we find it is practical to explore a novel setting of active finetuning using \textit{annotated} data. 
Then, we aim to select an optimal subset of training data for finetuning while having faster convergence and/or higher performance metrics as shown in Figure \ref{fig:teaser}. The selection can further be performed in a loop to accommodate the changing model performance during the finetuning process.
To this end, we propose to perform an Objective-aware Data Selection (ODS) using a parameterized data selection model. This ODS model reweighs training data distribution according to the downstream objective and selects a subset that is both diverse and representative to the task.

\begin{figure*}[t]
\includegraphics[width=1\textwidth]{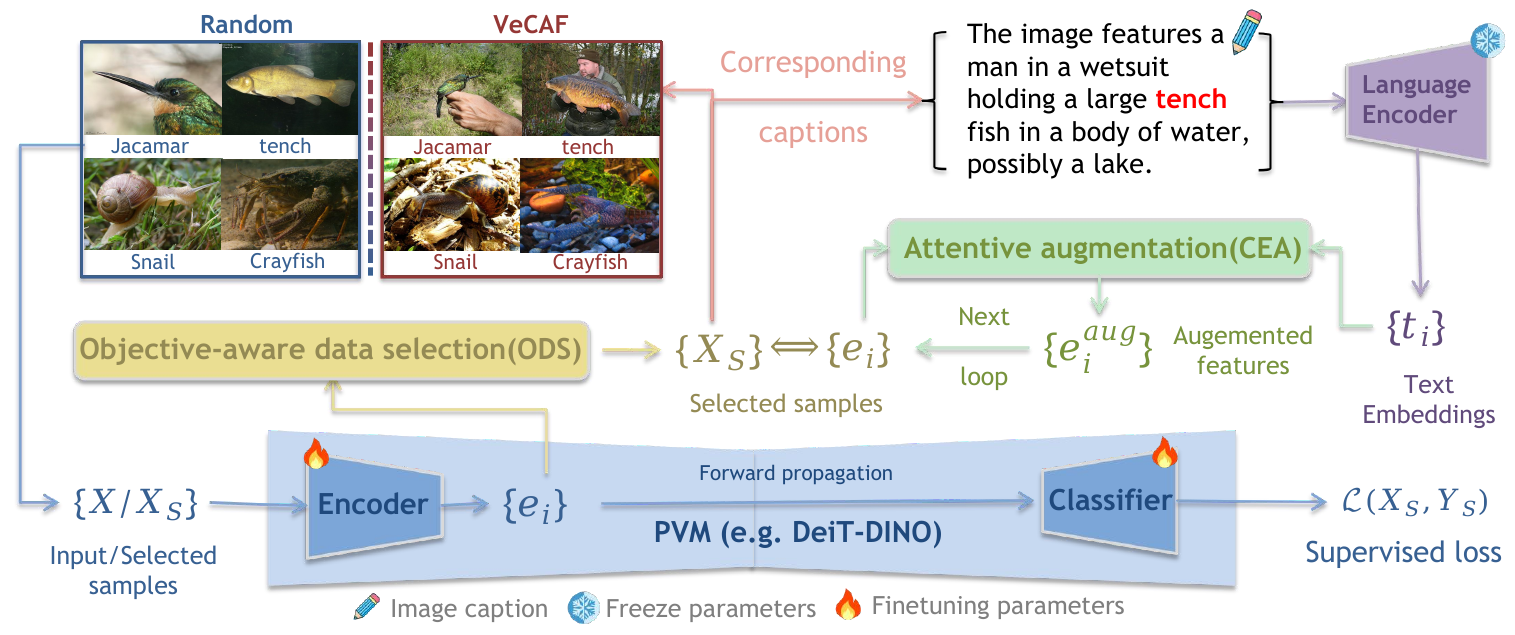}
\centering
\caption{\textbf{The overall framework of VeCAF.} In each data selection loop, VeCAF performs an Objective-aware Data Selection (ODS) to select more informative images for finetuning. Cross-attentive Embedding Augmentation (CEA) is performed on the selected images to further enrich the semantic information captured by the image embeddings by incorporating language knowledge of the caption.}
\label{fig:framework}
\end{figure*}

The pursuit of objective-awareness brings new challenges to the data selection. Intuitively, images with misleading object appearances and complicated backgrounds, as illustrated in Figure \ref{fig:framework}, often provide more informative supervision. However, the image features extracted by the PVM may not fully capture all the semantic information present in the image. Therefore, PVM image features may miss useful information for the data selection and finetuning process as in previous ActiveFT~\cite{xie2023active} work. 
To address this limitation, we propose to leverage semantically rich language embedding spaces of the text encoders (e.g., CLIP~\cite{radford2021learning}, mT5~\cite{xue2020mt5}, BERT~\cite{devlin2018bert} etc.) in our novel \textbf{V}ision-languag\textbf{e} \textbf{C}ollaborative \textbf{A}ctive \textbf{F}ine-tuning (VeCAF) approach.
Specifically, we extract the text embeddings of the captions associated with each image. These captions may be sourced directly from original datasets such as COCO-Caption or alternatively generated by a multimodal Large Language Model (LLM) e.g., BLIP-2~\cite{li2023blip}. Then, we propose a Cross-attentive Embedding Augmentation (CEA) to augment the image features extracted by the PVM such that the augmented features can focus more on the rich semantic information of the training samples. Therefore, the CEA facilitates both the active data selection and the finetuning.

Empirically, we demonstrate improved efficiency and performance across different scenarios.
VeCAF is evaluated on three image classification datasets including CIFAR-10~\cite{krizhevsky2009learning}, Caltech101~\cite{fei2004learning}, and ImageNet-1K~\cite{deng2009imagenet} with the pretraining-finetuning paradigm where base models are pretrained on ImageNet-1K.
Results on ImageNet demonstrate that VeCAF can significantly accelerate the PVM convergence speed to the target performance and saves up to 3.3$\times$ computational cost when compared to finetuning with all training set.
Importantly, VeCAF also addresses out-of-distribution (OOD) scenarios, where we augment image features by target-domain text embeddings derived from the generated image captions. By leveraging the alignment between text and images embeddings, VeCAF increases the likelihood of selecting images that possess the characteristic features of the target domain from the training dataset as shown in Figure \ref{fig:da}. In addition, we verify the OOD generalization ability on the corrupted ImageNet-C dataset where VeCAF improves accuracy by over 6\% when compared to state-of-the-art active learning methods.
Our main contributions are summarized as follows:
\begin{itemize}
\setlength\itemsep{0em}
    \item We propose a novel framework, VeCAF, to improve computational efficiency of PVM finetuning using both the training objective and the language-embedded knowledge.
    \item We propose the Objective-aware Data Selection (ODS), where a parameterized data selection model is optimized for the user-specified objectives and selects a subset that contributes to faster convergence and higher performance metrics.
    \item We further employ pretrained language encoders with the proposed Cross-attentive Embedding Augmentation (CEA) to enrich semantic information in image features and to provide explicit semantic guidance for data selection and finetuning.
\end{itemize}


\section{Related work}

\subsection{Active learning}
The learning algorithm in active learning is allowed to choose the data from which it learns~\cite{settles2009active}. 
There are two main selection criteria: uncertainty~\cite{beluch2018power,yuan2021multiple,liu2021influence,zhang2024efficient} and diversity~\cite{agarwal2020contextual,wang2023large,cai2023human,sener2017active}. Uncertainty of the model can aid selection of the most difficult unlabeled data. Early works estimate the uncertainty with various heuristics such as posterior probability~\cite{lewis1994heterogeneous,wang2016cost,zhang2023unimodal,liu2022multi}, entropy~\cite{joshi2009multi,luo2013latent} and classification margin~\cite{tong2001support}. 
Previous works~\cite{mahmood2021low,sener2017active} also formulate active learning as an optimization problem. They typically operate in a discrete space that trivially matches the sample distribution of a dataset~\cite{eric2007active,jha2022bayesian}. However, discrete optimization is harder to solve than the one in continuous space. 
Also, most previous methods are designed for a from-scratch training without the pretraining stage. Bengar et al.~\cite{bengar2021reducing} reveals drawbacks of such setting without unsupervised pretraining. Xie et al.~\cite{xie2023active} addresses both shortcomings with the proposed continuous-space ActiveFT method that applies selected samples to the finetuning of the
pretrained model in a single pass. VeCAF extends ActiveFT to the practical setting of data selection from a large labeled dataset. VeCAF framework enhances the training efficiency with a training objective-aware data selection and achieves optimal finetuning results with minimal training batches.


\subsection{Exploiting language in vision model training}

Recent developments in language models, especially vision-language models~\cite{li2023blip,li2023m} demonstrate their effectiveness in aligning the embedding space of vision and language to achieve cross-modal generalization.
For example, BLIP-2~\cite{li2023blip} addresses the modality gap using a lightweight Querying Transformer, while Shikra~\cite{chen2023shikra} handles spatial coordinate inputs and outputs in natural language and excels in referential dialogue and general vision-language tasks.
Many prior work exploit the connection between the image and text modalities, where they explore the use of language in training better vision models. For example, Ma et al.~\cite{ma2023borrowing} leverage pretrained language models to design a distribution alignment objective. This objective guides the vision model to learn linguistic representations specific to the task under a semi-supervised setting. Similarly, Fahes et al.~\cite{fahes2023poda} utilize CLIP to optimize affine transformations of source domain features. This optimization aligns these features with the target text embeddings while preserving their content and semantics.
In our work, we use the language embeddings of image captions to perform text-space augmentations, achieving better sample selection quality in active learning setting.

\section{Method}
This section introduces the details of the proposed VeCAF framework. As illustrated in Figure \ref{fig:framework}, we start by selecting a subset of training data with the PVM image feature space using the proposed ODS method, as introduced in Section \ref{3.1}. Then, the selected samples pass through pretrained language model to get semantically-rich text embeddings, which augment the image features with our proposed CEA technique, as formulated in Section \ref{3.3}. The augmented image features are used for PVM finetuning as candidates in future rounds of data selection. Finally, we show that VeCAF can overcome challenges of out-of-distribution data in Section \ref{3.5}.

\subsection{Training objective-aware data selection}
\label{3.1}

Consider a PVM $f(\cdot;w)$ with $w$ weights and a user-defined finetuning objective $\mathcal{L}$. Given a labeled training set $\mathcal{D}: \{\mathcal{X,Y}\}$, where $\mathcal{X}$ is the set of image $x$ and $\mathcal{Y}$ is the set of corresponding label $y$. The goal of our active data selection is to find a subset $\mathcal{S}: \{\mathcal{X}_S,\mathcal{Y}_S\} \subset \mathcal{D}$, such that PVM finetuning with this subset for a fixed number of iterations leads to the largest reduction of the training objective. Formally, we formulate the optimization problem of our objective-aware data selection (ODS) process as
\begin{equation}
\mathcal{S}_{opt}= \arg\min_{\mathcal{S}} \mathbb{E}_{x,y\in\mathcal{D}}\left[\mathcal{L}\left(f(x;w-\beta\delta_{\mathcal{S}}),y\right)\right],
\label{equ:opt}
\end{equation}
where $\delta_\mathcal{S} = \nabla_w \mathbb{E}_{x_s,y_s\in\mathcal{S}}\left[\mathcal{L}\left(f(x_s;w),y_s\right)\right]$ is the gradient accumulated by finetuning with $\mathcal{S}$ and $\beta$ is the learning rate. 

In a practical setting, we cannot compute the gradient $\nabla_w\mathcal{L}\left(f(x;w),y\right)$ for each training example $(x,y)$ before the data selection, as the gradient computation cost almost equals to the cost of full finetuning, which contradicts the purpose of active data selection. In this sense, we make an assumption that a data point with a larger loss contributes more to the convergence speed of the model. 

Intuitively, this assumption leads a naive data selection policy of selecting the data points with Top-$K$ training losses. However, previous active learning work~\cite{xie2023active} has discovered that the diversity of the selected data is also important in order to cover corner cases in the dataset and to avoid overfitting. Therefore, we design the ODS algorithm with the following principles:
\textbf{1)} \textit{data point with a larger loss $\mathcal{L}\left(f(x;w),y\right)$ shall be selected with a higher probability}; and 
\textbf{2)} \textit{maintaining the diversity of data points selected in $\mathcal{S}$}. Analytically, we formulate our ODS objective as
\begin{equation}
    \mathcal{S}_{opt}= \arg\min_{\mathcal{S}} D_{KL}(p_{\mathcal{L}}(\mathcal{D})||p_S(\mathcal{S})) - \lambda R(p_S(\mathcal{S})),
\label{equ:ODS}
\end{equation}
where $D_{KL}(\cdot||\cdot)$ is the KL divergence, $R(\cdot)$ is a diversity metric, and $\lambda$ is a tradeoff factor.
$p_{\mathcal{L}}(\mathcal{D})$ is the distribution of the full training set that guides data selection. To follow our first principle, we assign the probability of each data point $(x,y)$ in $p_{\mathcal{L}}(\mathcal{D})$ according to the finetuning objective $\mathcal{L}$ scaled by a $Z$ normalization factor as 
\begin{equation}
    p_\mathcal{L}(x,y) = \mathcal{L}\left(f(x;w),y\right)/Z.
\label{equ:pd}
\end{equation} 

The distribution of the selected data  $p_S(\mathcal{S})$ is determined by the data selection model. 
As a sanity check, the naive ``Top-$K$ training losses'' serves as the optimal solution for Equation \ref{equ:ODS} without considering diversity when $\lambda=0$. 

To enable continuous optimization, we optimize Equation \ref{equ:ODS} using a parameterized data selection model $\theta_S$. For the simplicity, we follow previous work~\cite{xie2023active} to model the data selection distribution $p_S(\mathcal{S})$ in the lower-dimension image embedding space, where embedding $e$ is produced by the PVM from the input $x$ at a hidden layer. $\theta_S$ consists of $K$ centroids in the image embedding space, each selecting the nearest data point. We define the probability of a data point $x_i$ with the corresponding embedding $e_i$ being selected as
\begin{equation}
    p_S(x_i) = \exp(\langle e_i,\theta_S^{c_i} \rangle) / \sum\nolimits_{x_j\in\mathcal{D}}\exp(\langle e_j,\theta_S^{c_j} \rangle),
\label{equ:ps}
\end{equation}
where $\langle \cdot,\cdot \rangle$ denotes the cosine distance, and $\theta_S^{c_i}$ is the closest centroid in $\theta_S$ to $e_i$. We derive the parameterized distribution distance $D(\theta_S):=D_{KL}(p_\mathcal{L}||p_S)$ for optimization objective Equation \ref{equ:ODS} using Equation \ref{equ:pd} and \ref{equ:ps} as

\begin{equation}
\begin{aligned}
D(\theta_S) &= \sum\nolimits_{(x_i,y_i)\in\mathcal{D}} p_\mathcal{L}(x_i,y_i) \log \frac{p_\mathcal{L}(x_i,y_i)}{p_S(x_i)} \\
&= \mathbb{E}_{p_\mathcal{L}}[\log p_\mathcal{L}(x_i,y_i)] - \mathbb{E}_{p_\mathcal{L}}[\log p_S(x_i)] \\
&= C - \alpha \sum_{(x_i,y_i)\in\mathcal{D}} \mathcal{L}\left(f(x_i;w),y_i\right) \langle e_i,\theta_S^{c_i} \rangle,
\label{equ:dist}
\end{aligned}
\end{equation}
where $C$ and $\alpha$ are the constants omitted in the formulation.

For the diversity metric $R(\cdot)$, we follow the diversity regularization term proposed in~\cite{xie2023active} as
\begin{equation}
    R(\theta_S) = -\sum\nolimits_{\theta_S^i}\left[\log\sum\nolimits_{\theta_S^j, j\neq i} \exp(\langle\theta_S^i, \theta_S^j\rangle)\right].
\label{equ:diver}
\end{equation}

By substituting Equation \ref{equ:dist} and Equation \ref{equ:diver} into Equation \ref{equ:ODS} and by removing constant terms, our final objective in terms of $\theta_S$ parameters can be written as
\begin{equation}
\begin{aligned}
\theta_S^* = \arg\min_{\theta_S} &-\sum_{(x_i,y_i)\in\mathcal{D}} \mathcal{L}\left(f(x_i;w),y_i\right) \langle e_i,\theta_S^{c_i} \rangle\\
&+ \lambda\sum_{\theta_S^i}\left[\log\sum_{\theta_S^j, j\neq i} \exp(\langle\theta_S^i, \theta_S^j\rangle)\right].
\label{equ:objective}
\end{aligned}
\end{equation}

The optimization on $\theta_S$ is conducted via gradient descent. To further resolve the dependency of the optimized data selection model to its initialization as observed in~\cite{xie2023active}, we consider an independently-initialized data selection model ensemble, denoted as $\{\theta_{e}\}_{e=1}^{E}$, in the optimization. We empirically set $E=5$ to balance the performance-cost tradeoff.
After optimizing each data selection model $\theta_e$ independently using Equation \ref{equ:objective}, we can then remove the bias in the initialization by utilizing the mean $\mu$ and covariance $\Sigma$ calculated on $\theta_e^*$. Specifically, given an optimized data selection model $\theta_1^*$, we achieve the final unbiased data selection model as $\theta_{S}^* = \theta_{1}^*-\Sigma^{-1}(\theta_{1}^*-\mu)$.

The optimization in Equation \ref{equ:objective} is performed before each finetuning ``loop'' with the current model weights $w$. The training data with the closest embedding to each $\theta_S$ centroid is selected to form a ``finetuning set'' with $K$ elements. Then, this set is used to finetune the PVM until the next loop starts. We update ODS weights $w$ after a predetermined number of training batches.

\subsection{Cross-attentive embedding augmentation}
\label{3.3}

We empirically find that samples selected by the ODS process tend to consist of multiple objects in both foreground and background, which is a result of coarse image embeddings produced by the PVM. To further improve quality of image embeddings for both data selection and finetuning, we propose Cross-attentive Embedding Augmentation (CEA). Given a training image, CEA leverages a text encoder to transform the image caption to the corresponding text embedding. Then, the text embedding is used to augment the image embedding with attention-based method.

Given selected sample $\{x_{S_{i}}\}_{i=1}^{K}$ in the previous ODS run, the corresponding image caption can be denoted as $\{c_{S_{i}}\}_{i=1}^{K}$. Then we feed the caption into a frozen text encoder, e.g. BERT~\cite{devlin2018bert}, to convert the captions into the text embeddings $\{t_{i}\}_{i=1}^{K}$. We use text embeddings with the same dimensions as image embeddings for easier fusion.

Inspired by~\cite{ji2019learning}, CEA is conducted as mapping the image embedding $e_i$ towards the corresponding text embedding $t_i$. To decide the magnitude of the augmentation, we compute a sample-wise attention score $\{\alpha_{i}\}_{i=1}^{K}$ with the cosine distance between $e_i$ and $t_i$ as
\begin{equation}
\begin{aligned}
    \alpha_{i} = \text{Softmax}(\frac{e_{j} \cdot t_{j}}{\|e_{j}\|_2 \; \|t_{j}\|_2}) = \frac{\exp\left(\frac{e_{i} \cdot t_{i}}{\|e_{i}\|_2 \; \|t_{i}\|_2}\right)}{\sum_{j=1}^{K}\exp\left(\frac{e_{j} \cdot t_{j}}{\|e_{j}\|_2 \; \|t_{j}\|_2}\right)}.
\label{attention}
\end{aligned}
\end{equation}

The attention score $\alpha_i$ helps to derive the augmented embedding $e^{aug}_{i}$ using the image embedding $e_i$ and the corresponding text embedding $t_i$ as
\begin{equation}
    e^{aug}_{i} = e_{i} - \eta \alpha_i (e_i - t_i),
\label{equ:aug}
\end{equation}
where $\eta$ is the fixed step size.
The proposed CEA method enriches the semantic information and improves performance after finetuning as shown in experiments. We further present the pseudo code of VeCAF in Algorithm 1, which specifies the complete procedure of the proposed Vision-language Collaborative Active Finetuning (VeCAF). 

\begin{algorithm}[t]
    \SetAlgoLined
    \SetKwInOut{Input}{input}
    \Input{Objective-aware data selection $ODS(\cdot;)$, labeled data pool $(\mathcal{X,Y})$, image caption pool $Cap$, PVM $f(\cdot;w)$, pretrained language encoder $LM(\cdot)$, data selection loop number $L$, batch number $B$ for each loop }
    \SetKwInOut{Output}{output}
    \Output{Finetuned vision model $f(\cdot;w_{FT})$}
    \For{loop $\in [L]$}{
        Obtained the selected sample pool $\mathcal{S}_{opt}=ODS((\mathcal{X,Y});f(\cdot;w))$ \;
        Get the corresponding image caption $Cap_{i}$ for $s_i\in \mathcal{S}_{opt}$ \;
        Transfer the image caption to text embedding as $t_{i}=LM(Cap_{i})$ \;
        \tcc{\rongyu{Cross-attentive embedding augmentation (CEA)}}
        CEA attention score computation $\alpha_{i} = \text{Softmax}(\frac{e_{j} \cdot t_{j}}{\|e_{j}\|_2 \; \|t_{j}\|_2})$ \;
        Image embedding $e_{i}$ augmentation $e^{aug}_{i} = e_{i} - \eta\cdot\alpha_i (e_i - t_i)$ \;
        \tcc{\rongyu{PVM finetuning with $\mathcal{S}_{opt}$}}
        \For{batch $\in[B]$}{
        Sample next batch from $\mathcal{S}_{opt}$ \;
        Calculate the loss with the classifier \;
        Optimize $f(\cdot;w_{FT})$ via gradient descent \;
        }
        $f(\cdot;w) \leftarrow f(\cdot;w_{FT})$
    }
    Return the finetuned vision model $f(\cdot;w_{FT})$
\caption{Vision-language Collaborative Active Finetuning (VeCAF)}
\label{pc:vecaf}
\end{algorithm}

\begin{figure}[t]
\includegraphics[width=\linewidth]{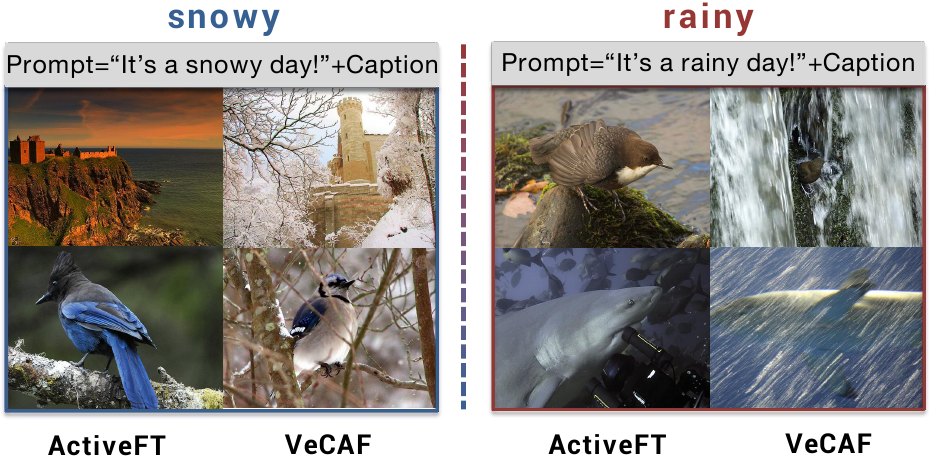}
\centering
\caption{The selected samples of ActiveFT~\cite{xie2023active} and VeCAF. With the caption augmented: ``\textit{It is a $\{snowy/rainy\}$ day!}'', VeCAF can select images that correspond to the target domain.}
\label{fig:da}
\end{figure}

\subsection{Improving out-of-distribution scenarios}
\label{3.5}
In addition to achieving more efficient finetuning on in-distribution (ID) tasks, one of the additional benefits of introducing the text modality is the ability to artificially modify the corresponding image captions as a semantic augmentation. By leveraging the strong language capability provided by the text encoder, we can implicitly alter our requirements for the selected data. This capability enables us to facilitate domain transfer to out-of-distribution (OOD) scenarios with only ID data.

For example, leveraging the CEA technique, we can add descriptive phrases like ``\textit{It's a \{target\_domain\} day!}'' to the image captions. This modification shifts the PVM image embedding towards the ``snow'' distribution while highlighting the ID sample images that have snowflakes or similar patterns during the data selection process, as demonstrated in Figure \ref{fig:da}. By incorporating such textual cues, we can guide the selection process toward images that possess specific characteristics or attributes, so the finetuned model can better generalize in out-of-distribution scenarios.

\setlength\tabcolsep{3pt}
\begin{table*}[h]
\centering
\caption{\textbf{Classification accuracy with the fixed training cost.} All methods for each dataset are trained using the fixed number of batches. The percentage value on top reports the ratio of data selected during each loop. Some results marked as N/A (``-'') as explained in Appendix B. Top-1 accuracy with a standard error of 3 repetitions is reported, \%. }
\resizebox{\linewidth}{!}{%
\begin{tabular}{c|c|ccc|ccc|ccc}
  \toprule
             \multirow{2}*{Method} &  \multirow{2}*{Loop}    & \multicolumn{3}{c|}{CIFAR-10} & \multicolumn{3}{c|}{Caltech101}  & \multicolumn{3}{c}{ImageNet-1K}\\
             \multirow{2}{*}{} & \multirow{2}{*}{} &
            1\% & 2\% & 5\% & 2\% & 5\% & 10\% & 1\% & 2\% & 4\%   \\
\midrule
            \multirow{2}*{LearnLoss\cite{yoo2019learning}} & single-run & 85.93\tiny$\pm$0.05 & 91.22\tiny$\pm$0.08 & 93.89\tiny$\pm$0.07 & 46.47\tiny$\pm$0.16 & 43.74\tiny$\pm$0.13 & 65.59\tiny$\pm$0.05 & 49.37\tiny$\pm$0.09 & 57.86\tiny$\pm$0.07 & 63.46\tiny$\pm$0.11 \\
            \multirow{2}{*}{} & multi-run & 87.53\tiny$\pm$0.14 & 92.53\tiny$\pm$0.17 & 94.43\tiny$\pm$0.22 & 47.36\tiny$\pm$0.13 & 44.27\tiny$\pm$0.19 & 66.27\tiny$\pm$0.14 & 49.89\tiny$\pm$0.16 & 58.22\tiny$\pm$0.17 & 64.18\tiny$\pm$0.13 \\
\midrule 
            \multirow{2}*{TA-VAAL\cite{kim2021task}} & single-run & 85.46\tiny$\pm$0.10 & 92.65\tiny$\pm$0.06 & 94.85\tiny$\pm$0.10 & 59.26\tiny$\pm$0.05 & 58.11\tiny$\pm$0.08 & 66.94\tiny$\pm$0.08 & - & - & 63.86\tiny$\pm$0.13 \\
            \multirow{2}{*}{} & multi-run & 87.74\tiny$\pm$0.17 & 93.77\tiny$\pm$0.19 & 96.01\tiny$\pm$0.12 & 60.57\tiny$\pm$0.18 & 59.25\tiny$\pm$0.15 & 67.32\tiny$\pm$0.21 & - & - & 64.32\tiny$\pm$0.16 \\
\midrule 
            \multirow{2}*{ALFA-Mix\cite{parvaneh2022active}} & single-run & 86.69\tiny$\pm$0.07 & 92.87\tiny$\pm$0.06 & 95.14\tiny$\pm$0.08 & 59.73\tiny$\pm$0.05 & 58.74\tiny$\pm$0.09 & 67.36\tiny$\pm$0.08 & - & - & 64.03\tiny$\pm$0.07 \\
            \multirow{2}{*}{} & multi-run & 88.14\tiny$\pm$0.13 & 93.26\tiny$\pm$0.13 & 95.75\tiny$\pm$0.11 & 60.74\tiny$\pm$0.16 & 60.47\tiny$\pm$0.14 & 68.25\tiny$\pm$0.15 & - & - & 64.69\tiny$\pm$0.19 \\
\midrule 
            \multirow{2}*{ActiveFT~\cite{xie2023active}} & single-run & 90.91\tiny$\pm$0.12 & 93.80\tiny$\pm$0.09 & 95.39\tiny$\pm$0.08 & 62.86\tiny$\pm$0.05 & 60.55\tiny$\pm$0.09 & 69.34\tiny$\pm$0.06 & 53.96\tiny$\pm$0.07 & 60.33\tiny$\pm$0.09 & 64.72\tiny$\pm$0.10\\
            \multirow{2}{*}{} & multi-run & 92.79\tiny$\pm$0.11 & 94.17\tiny$\pm$0.16 & 95.92\tiny$\pm$0.14 & 64.43\tiny$\pm$0.12 & 61.97\tiny$\pm$0.17 & 71.42\tiny$\pm$0.14 & 55.67\tiny$\pm$0.13 & 61.86\tiny$\pm$0.21 & 65.18\tiny$\pm$0.17\\
\midrule
            Full Data FT & single-run & \multicolumn{3}{c|}{93.64\tiny$\pm$0.02} & \multicolumn{3}{c|}{62.08\tiny$\pm$0.02} & \multicolumn{3}{c}{57.53\tiny$\pm$0.01} \\
\midrule
            VeCAF(ours) & multi-run & \textbf{93.57}\tiny$\pm$0.02 & \textbf{95.27}\tiny$\pm$0.04 & \textbf{96.24}\tiny$\pm$0.02 & \textbf{66.33}\tiny$\pm$0.04 & \textbf{65.15}\tiny$\pm$0.03 & \textbf{72.21}\tiny$\pm$0.03 & \textbf{58.31}\tiny$\pm$0.04 & \textbf{63.76}\tiny$\pm$0.03 & \textbf{66.57}\tiny$\pm$0.02\\
\bottomrule
		\end{tabular}
	}
	\label{main_result}
\end{table*}

\section{Experiments}
\label{section:experiment}

In this section, we explain our experiment protocols and conduct experiments on multiple image classification tasks. We demonstrate the superiority of the proposed VeCAF in both in-distribution and out-of-distribution scenarios and improve the finetuning efficiency by up to 3.3$\times$ and 2.7\% on ImageNet-1K. 
We first introduce the experiment setup in Sec.~\ref{subsection:setup}.
Main results in Sec.~\ref{subsection:main_results} show the effectiveness and efficiency of the proposed VeCAF compared with other sample selection methods.
In Sec.~\ref{subsection:ablation_study}, we present further analysis from different perspectives.

\subsection{Experiment setup}
\label{subsection:setup}

\paragraph{Datasets}
For model training, we conduct experiments using three image classification datasets: CIFAR-10~\cite{krizhevsky2009learning}, class-unbalanced Caltech101~\cite{fei2004learning}, and ImageNet-1K~\cite{deng2009imagenet}. Details of the dataset can be found in Appendix A. For out-of-distribution evaluation, we evaluate on ImageNet-C~\cite{hendrycks2019benchmarking} which consists of synthetically generated corruptions applied to the ImageNet validation set.

\paragraph{Implementation details}
In our main experiments, we use the DeiT-B model~\cite{pmlr-v139-touvron21a} pretrained with DINO~\cite{caron2021emerging} on ImageNet-1K~\cite{deng2009imagenet} as the PVM for finetuning. Additionally, we present experiments for different PVM architectures and model sizes in Section \ref{subsection:ablation_study} to demonstrate VeCAF generalizability.
For all experiments, we resize input images to 224$\times$224 to ensure consistency during both the data selection and the finetuning. In the ODS process, we optimize the data selection model parameters $\theta$ using the Adam optimizer with a learning rate of 0.001 until convergence. We follow~\cite{pmlr-v139-touvron21a} to finetune the PVM with the selected data.

We adopt the standard protocols outlined in \cite{pmlr-v139-touvron21a} to finetune the DeiT-Base model. For the CIFAR-10~\cite{krizhevsky2009learning} and the Caltech101~\cite{fei2004learning}, we perform supervised finetuning of the pretrained models using the SGD optimizer. We set the learning rate (lr) to 5e-4, weight decay to 1e-4, and momentum to 0.9. The total batch number used for training CIFAR-10 and Caltech101 is 750 and 1500, respectively.
For the ImageNet-1K~\cite{deng2009imagenet}, the SGD optimizer with the same hyperparameters as CIFAR-10 and Caltech101 is employed. The total number of used batches for training is 125000.
These experiments are conducted on two Tesla-A100 GPUs. Each GPU processes a batch size of 256, and we apply cosine learning rate decay on selected subsets of the training data. 

\paragraph{Evaluation protocol}
We primarily focus on an efficient training setting, where all the  algorithms are only allowed to train with the same batch size and the same number of batches. Convergence results with unlimited batches are in Appendix C.3. As a default setting, we perform multi-run data selection before each loop with the fixed training batches for all baselines including VeCAF. Three loops are used in the main experiments, where each loop is defined as 1/3-rd of the total number of training batches.

\paragraph{Baselines}
We compare VeCAF with three active learning baselines LearnLoss~\cite{yoo2019learning}, TA-VAAL~\cite{kim2021task}, ALFA-Mix~\cite{parvaneh2022active}, ActiveFT~\cite{xie2023active} and conventional full data finetuning. 
\begin{itemize}
   \item LearnLoss~\cite{yoo2019learning} predicts target losses for unlabeled inputs to identify potential incorrect predictions.
   \item TA-VAAL~\cite{kim2021task} considers the data distribution of labeled and unlabeled pools and enhances the learning process by incorporating a ranking loss prediction.
   \item ALFA-Mix~\cite{parvaneh2022active} employs interpolations between labeled and unlabeled instances to uncover unrecognized features, which derives an efficient implementation using a closed-form solution to identify the optimal interpolation that induces changes in predictions.
   \item ActiveFT~\cite{xie2023active} focuses on selecting the data subset that exhibits similar distribution to the unlabeled pool, while maintaining the diversity within the selected subset by optimizing a parametric model in the continuous space.
   \item Full Data FT takes all the data from the training set to finetune the pretrained vision models.
\end{itemize}

\setlength\tabcolsep{4pt}
\begin{table}[t]
	\centering
 	\caption{\textbf{OOD generalization ability.} Models are trained with 1\% data subset per loop using the uncorrupted ImageNet. Evaluation results are for the distorted ImageNet-C validation set. }
	\resizebox{0.85\columnwidth}{!}{
		\begin{tabular}{cccc}
  \toprule
            Source & Target (eval.)  & Method & Top-1 Acc. \\ 
\midrule
		\multirow{10}{*}{\begin{tabular}[c]{@{}c@{}}ImageNet\end{tabular}}    
             & \multicolumn{3}{c}{\cellcolor{lightgray}Prompt: It's a snowy day.}  \\ 
            \multirow{10}{*}{}   & \multirow{4}{*}{\begin{tabular}[c]{@{}c@{}}ImageNet-C\\Snowy\end{tabular}}  & Source-only & 38.89\tiny$\pm$0.20  \\
            \multirow{10}{*}{}   &   & CLIPStyler &  40.71\tiny$\pm$0.41 \\
            \multirow{10}{*}{}   &  & ActiveFT &  36.36\tiny$\pm$0.11 \\
            \multirow{10}{*}{}   &  & VeCAF &  \textbf{42.33}\tiny$\pm$0.03 \\
		  \multirow{10}{*}{} 
             & \multicolumn{3}{c}{\cellcolor{lightgray}Prompt: It's a foggy day.}  \\ 
            \multirow{10}{*}{}   & \multirow{4}{*}{\begin{tabular}[c]{@{}c@{}}ImageNet-C\\Foggy\end{tabular}}  & Source-only & 45.55\tiny$\pm$0.27  \\
            \multirow{10}{*}{}   &   & CLIPStyler &  46.25\tiny$\pm$0.36 \\
            \multirow{10}{*}{}   &  & ActiveFT &  42.45\tiny$\pm$0.08 \\
            \multirow{10}{*}{}   &  & VeCAF &  \textbf{47.71}\tiny$\pm$0.03 \\
\bottomrule
		\end{tabular}
  }
	\label{ood}
\end{table}

\subsection{Main results}
\label{subsection:main_results}
\paragraph{In-distribution results}
\textit{Overall performance:} 
We present the image classification results in Table \ref{main_result}. The results demonstrate the superior effectiveness of the proposed VeCAF method when compared to other active learning approaches. For a fair comparison, we have also extend the four active learning baselines to a multi-run framework. This means that we apply these methods to select the data finetuning subset at the beginning of each loop using different random seeds. We can observe from the results that traditional active learning methods often encounter challenges within the pretraining-finetuning paradigm, which aligns with our findings. In contrast, VeCAF consistently outperforms other methods across all three datasets, irrespective of the employed sampling ratios. Remarkably, even with low sampling ratios, our method excels in selecting highly representative samples. When compared to full data finetuning with limited training batches, VeCAF achieves higher accuracy even with only 1\% of the data being used for finetuning in each epoch. This practical advantage is significant as it allows for supervised finetuning with a smaller number of samples compared to the overall pool size with less steps, thereby reducing training costs. Additionally, it is noteworthy that VeCAF exhibits a more pronounced performance enhancement on more complex datasets, with an increase of over 2.7\% accuracy on ImageNet-1K compared to 1\% on CIFAR-10. 

\begin{figure}[t]
\includegraphics[width=\linewidth]{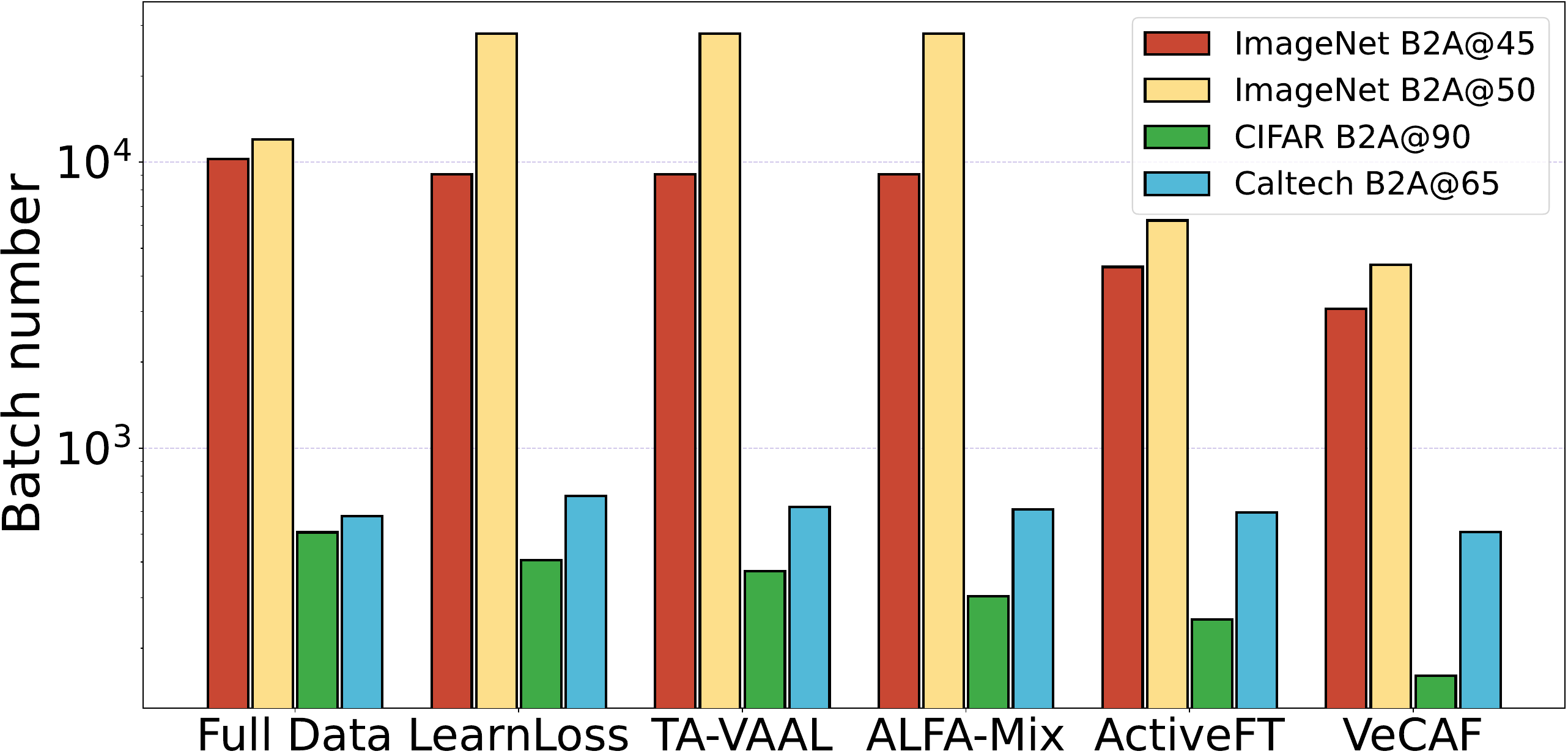}
\centering
\caption{\textbf{Comparison of training efficiency.} VeCAF requires significantly fewer training batches to reach the target accuracy (B2A) compared with other baselines and full-data finetuning. Note that the y-axis has an exponential scale.}
\label{fig:batch}
\end{figure}

\begin{figure*}[t]
	\centering
	\begin{minipage}{0.49\linewidth}
		\centering
		\includegraphics[width=0.9\linewidth]{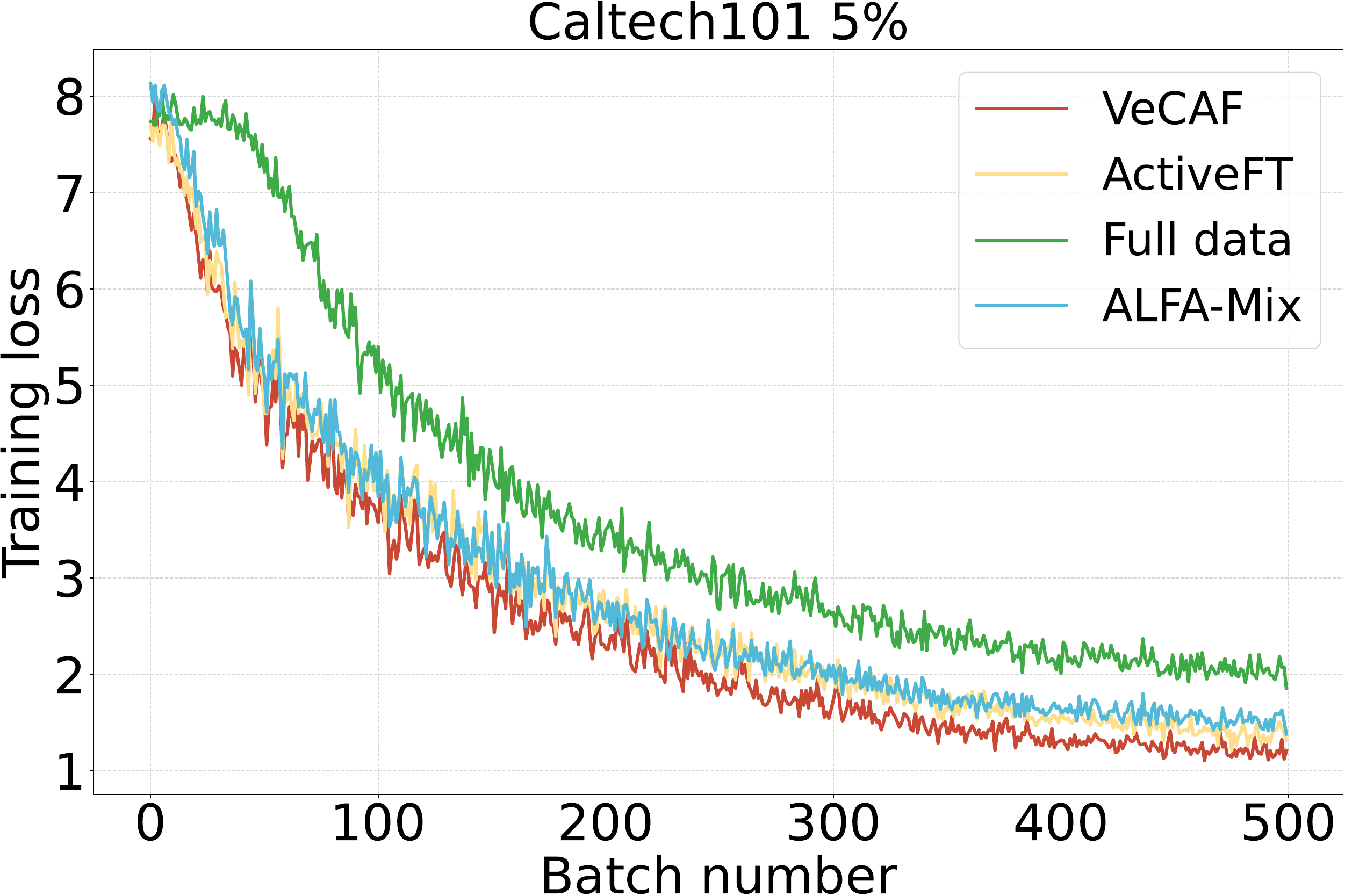}
	\end{minipage}
	\begin{minipage}{0.49\linewidth}
		\centering
		\includegraphics[width=0.9\linewidth]{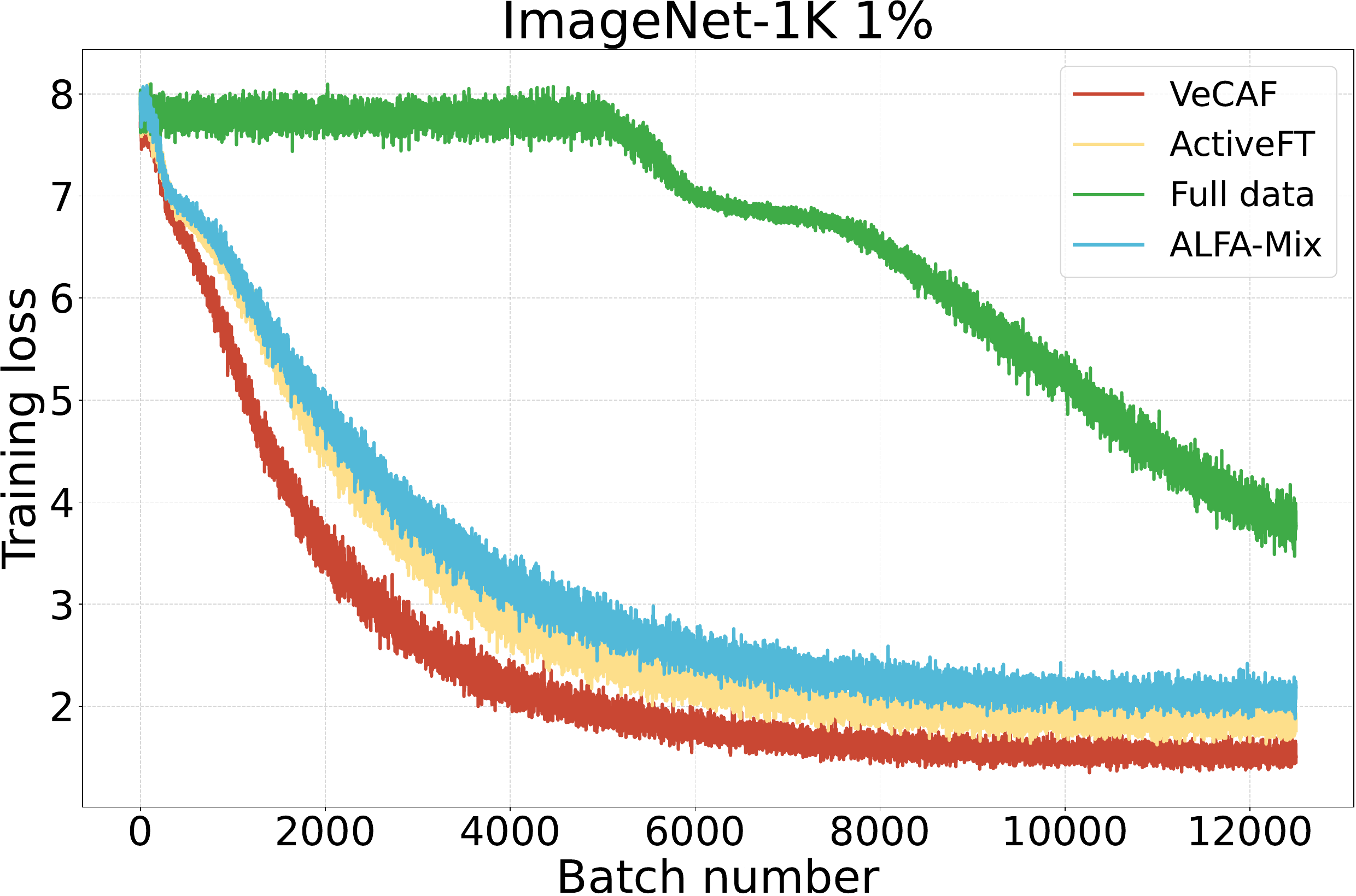}
	\end{minipage}
 		\caption{Training loss curve of VeCAF and other baselines including ActiveFT, ALFA-Mix, and Full data FT on Caltech-101 (left) and ImageNet-1K (right) with 5\% and 1\% data, respectively.}
   \label{loss}
\end{figure*}

\textit{Efficiency enhancement:} To provide a deeper understanding of the effectiveness of our proposed method, we have included figures to illustrate the required batch numbers to achieve the target accuracy (B2A) for different approaches on the three datasets. Specifically, we report training batches needed for each method with \textbf{B2A@45,50} (\textit{i.e., 50\% Top-1 accuracy}) for \textit{ImageNet}, \textbf{B2A@65} for \textit{Caltech101}, and \textbf{B2A@90} for \textit{CIFAR-10}, respectively. Figure \ref{fig:batch} displays such batch numbers, highlighting the efficiency of VeCAF in comparison to other methods. On ImageNet, VeCAF achieves 3.3$\times$ acceleration over full data finetuning (3075 \textit{v.s.} 10250 batches) and outperforms other baselines more significantly as the target accuracy goes higher.
Additionally, we plot the training loss as a function of the batch number for different approaches during the finetuning process in Figure \ref{loss}. This visualization further highlights the convergence speed and performance of VeCAF compared to alternative methods.
These figures provide a comprehensive view of the performance and efficiency of VeCAF, emphasizing its effectiveness in terms of training speed and achieving the desired accuracy levels.

\setlength\tabcolsep{3pt}
\begin{table}[t]
\footnotesize
	\centering
 	\caption{\textbf{1-vs.-all accuracy for certain categories on CIFAR-10.} DINO-S~\cite{kim2021task} is used as PVM with 2\% of data in each loop.}
	\resizebox{\columnwidth}{!}{
		\begin{tabular}{c|cccc}
  \toprule
            Methods & Airplane & Bird & Cat & Deer  \\
\midrule
            ActiveFT\cite{xie2023active}   & 89.50\tiny$\pm$0.01 & 82.70\tiny$\pm$0.03 & 83.00\tiny$\pm$0.02 & 87.50\tiny$\pm$0.04 \\ 
            Top-K Loss & 82.18\tiny$\pm$0.05 & 81.88\tiny$\pm$0.01 & 75.72\tiny$\pm$0.04 & 84.64\tiny$\pm$0.03 \\ 
            VeCAF(ours)  & \textbf{91.24}\tiny$\pm$0.02 & \textbf{88.80}\tiny$\pm$0.04 & \textbf{86.41}\tiny$\pm$0.02 & \textbf{89.12}\tiny$\pm$0.03 \\ 

\bottomrule
		\end{tabular}
  }
	\label{class_accuracy}
\end{table}

\textit{Finegrained training objective awareness:}
The proposed ODS method offers additional flexibility to accommodate finegrained training objective. For verification, we evaluate VeCAF under 1-vs.-all finetuning objective on multiple CIFAR-10 classes.
Specifically, given a target class, we set the loss as a binary classification, with target class being positive and all others being negative. The modified loss is used in ODS optimization as in Equation \ref{equ:objective}. 
A naive objective-aware baseline of selecting samples with the largest loss is also included in the comparison.
Table \ref{class_accuracy} shows that VeCAF accommodates the finegrained objective to better improve the performance of target categories.

\vspace{-0.5cm}
\paragraph{Out-of-distribution results} 
We also conduct experiments using ImageNet-C test set to assess its superiority under out-of-distribution (OOD) scenarios, as presented in Table \ref{ood}. Specifically, we consider the domain adaptation scenarios of clear$\to$snowy and clear$\to$foggy.

To provide a comprehensive evaluation, we compare VeCAF with the state-of-the-art baseline, CLIPstyler~\cite{kwon2022clipstyler}. Both VeCAF and CLIPstyler leverage the pretrained CLIP model and offer generalization capabilities for source images. However, it is essential to note that VeCAF uses active learning concept to select source-domain data points for finetuning, while CLIPstyler specializes in style transfer.
Furthermore, we compare the results obtained in the source-only setting, where models are finetuned exclusively on the source data, with full training set and ActiveFT~\cite{xie2023active} baselines. 1\% of the training data is used in each finetuning loop for ActiveFT and VeCAF.

In these experiments, we finetune the classifier using uncorrupted (source) ImageNet data points selected by ActiveFT and VeCAF, while CLIPstyler finetunes on stylized images. We can see from Table \ref{ood} that VeCAF consistently outperforms the three baselines by up to 4.2\%. CLIPstyler achieves better performance than other baselines, but multiple artifacts in the stylized images limit its performance. ActiveFT underperforms in the OOD setting with only 36.36\% and 42.45\% accuracy, respectively. This is a result of overfitting as ActiveFT selects data only based on the source domain distribution. On the other hand, VeCAF is superior in generalization ability, by leveraging a domain-specific text embedding augmentation on selection and finetuning enabled by the proposed CEA.


 

\setlength\tabcolsep{3pt}
\begin{table}[t]
\footnotesize
	\centering
 	\caption{Top-1 accuracy of VeCAF with different types and sizes of PVM backbones, and choices of pretrained language encoders. ImageNet-1K results are reported with 1\% data.}
	\resizebox{\columnwidth}{!}{
		\begin{tabular}{c|cccc}
  \toprule
            \diagbox{PVM}{LM} & CLIP & BERT-L & mT5-L & GPT2-L \\
\midrule
            DeiT-S   & 53.51\tiny$\pm$0.02 & 53.64\tiny$\pm$0.03 & 53.82\tiny$\pm$0.03 & 53.96\tiny$\pm$0.03 \\
            DeiT-B  & 58.31\tiny$\pm$0.04 & 58.63\tiny$\pm$0.04 & 58.71\tiny$\pm$0.03 & 58.87\tiny$\pm$0.01 \\
\midrule
            SwinT-S   & 53.66\tiny$\pm$0.03 & 53.71\tiny$\pm$0.03 & 53.89\tiny$\pm$0.02 & 54.01\tiny$\pm$0.02 \\
            SwinT-B  & 56.76\tiny$\pm$0.01 & 56.87\tiny$\pm$0.03 & 56.98\tiny$\pm$0.03 & 57.13\tiny$\pm$0.02 \\
\midrule
            XciT-M    & 58.48\tiny$\pm$0.04 & 58.70\tiny$\pm$0.03 & 58.77\tiny$\pm$0.03 & 58.95\tiny$\pm$0.03 \\
            XciT-L  & 61.13\tiny$\pm$0.01 & 61.37\tiny$\pm$0.01 & 61.56\tiny$\pm$0.03 & 61.78\tiny$\pm$0.03 \\
\bottomrule
		\end{tabular}
  }
	\label{vlm}
\end{table}

\subsection{Results analysis}
\paragraph{Generality of VeCAF}
The proposed VeCAF framework can be used to finetune various pretrained vision models (PVMs) with the help of different language encoders (LM). We apply it to DeiT-B~\cite{pmlr-v139-touvron21a} pretrained using the DINO framework~\cite{kim2021task} and several other PVMs such as DeiT-S, Swin-Transformer-S/B~\cite{liu2021swin}, and XciT-M/L~\cite{ali2021xcit}. 
Furthermore, we use other LMs including BERT-L~\cite{devlin2018bert}, mT5-L~\cite{xue2020mt5}, and GPT2-L~\cite{radford2019language} for augmentation by text embeddings. Table \ref{vlm} reports results on ImageNet-1K using different pairs of PVMs and LMs. VeCAF demonstrates its versatility and capability to adapt to different PVM model architectures, and the flexibility to be used with different text encoder models per user preferences. These results prove VeCAF to be a general active data selection method, that can incorporate the text embedding information from various configurations of LMs to improve the efficiency.


\setlength\tabcolsep{4pt}
\begin{table}[t]
	\centering
 \footnotesize
 	\caption{\textbf{Ablation study for the proposed techniques in VeCAF.} Data selection ratio is set to 1\% for CIFAR-10, 5\% for Caltech101, and 1\%  for ImageNet-1K in each loop.}
	\resizebox{\columnwidth}{!}{%
		\begin{tabular}{cc|ccc}
   \toprule
            ODS & CEA & CIFAR-10  &  Caltech-101 & ImageNet-1K     \\
    \midrule
             - & - & 92.79\tiny$\pm$0.12 & 60.55\tiny$\pm$0.10 & 55.67\tiny$\pm$0.13  \\
            \Checkmark & - & 93.27\tiny$\pm$0.03 & 64.11\tiny$\pm$0.04 & 57.73\tiny$\pm$0.04 \\
            - & \Checkmark & 93.15\tiny$\pm$0.05 & 63.04\tiny$\pm$0.06 & 56.13\tiny$\pm$0.05 \\
            \Checkmark & \Checkmark & \textbf{93.57}\tiny$\pm$0.02 & \textbf{65.15}\tiny$\pm$0.03 & \textbf{58.31}\tiny$\pm$0.04 \\
            \bottomrule
		\end{tabular}
	}
	\label{ablation}
\end{table}

\paragraph{Embedding visualization}
\label{visualization}

\begin{figure}[t]
	\centering
	\begin{minipage}{\linewidth}
		\centering
		\includegraphics[width=1\linewidth]{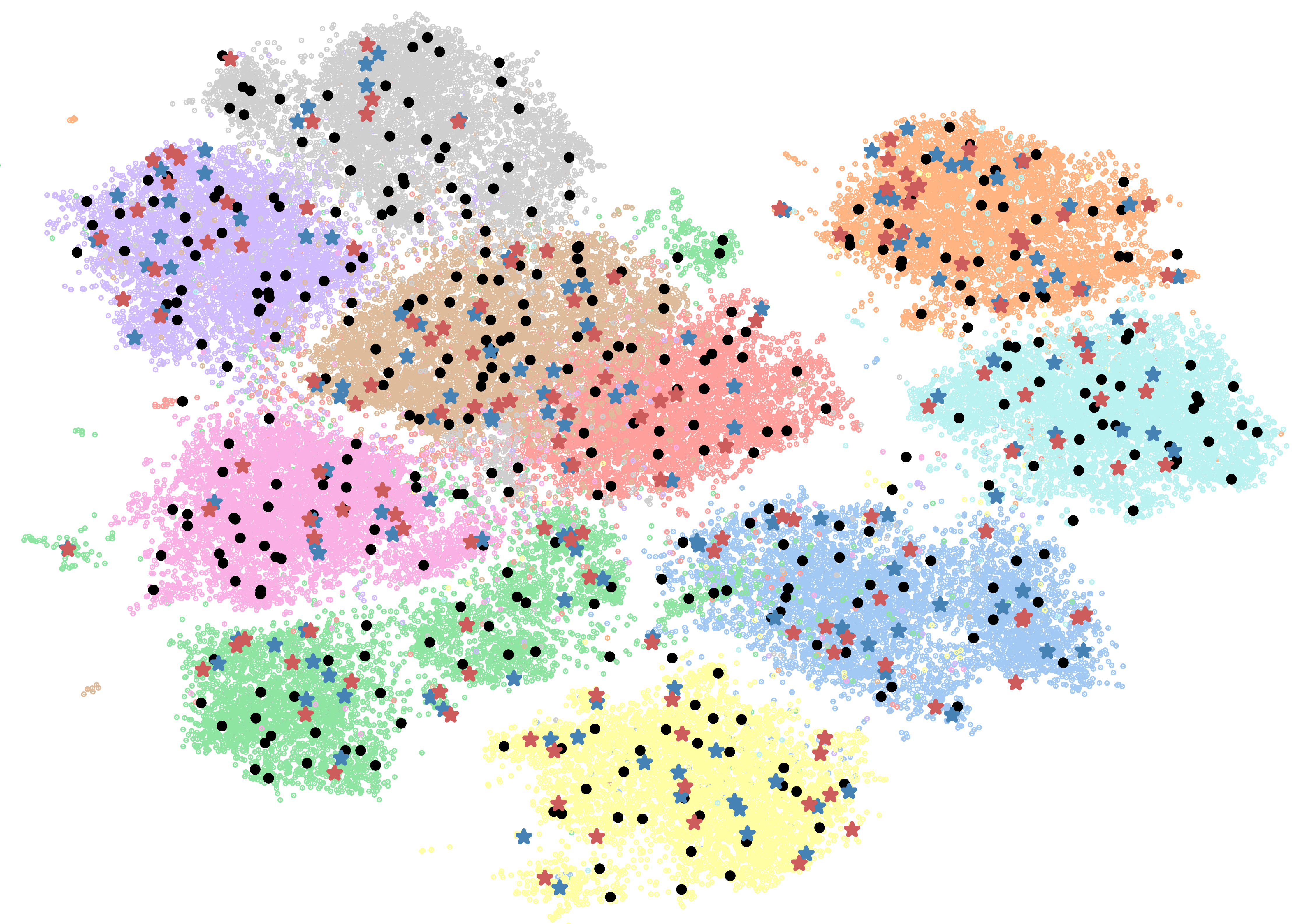}
		\caption{\textbf{UMAP visualization of training image embeddings.} Each background color represents one class. For selected samples, \textcolor{red}{$\bigstar$} suggests being selected by VeCAF only, \textcolor{blue}{$\bigstar$} suggests by ActiveFT~\cite{xie2023active} only, and $\bullet$ suggests by both.}
		\label{cifar}
	\end{minipage}
\end{figure}

In Table \ref{cifar}, we present the image embedding visualization of sample in the CIFAR-10 training set using UMAP dimension reduction.
With each method selecting 1\% of data from the training set, black dots represent the samples selected by both ActiveFT and VeCAF, red stars denote samples only selected by VeCAF, and blue stars denote samples only selected by ActiveFT. While maintaining the diversity of data selection, samples chosen by VeCAF appear to be closer to the boundaries compared to those selected by ActiveFT. This confirms that the proposed ODS helps to select important samples around the decision boundaries. This is a result of our selection strategy that incorporates training objective and, therefore, helps the PVM to learn the subtle differences between categories more efficiently in the finetuning.


\begin{table}[t]
	\centering
 	\caption{FLOPs and time for VeCAF, ActiveFT, and Full-FT}
	\resizebox{\columnwidth}{!}{
 \setlength\tabcolsep{1pt}
\begin{tabular}{c|cccccc} 
\toprule
    \multirow{2}*{Method} & Training & \multicolumn{4}{c}{Total FLOPs (G, all batches)}  & \multirow{2}*{Wallclock}  \\
    & batches  & CEA & ODS/DS & FT & ALL  & \\
\midrule
            VeCAF & 3000  & 9.53$\times 10^3$ & 1.5$\times 10^2$ & 2.11$\times 10^5$ & 2.21$\times 10^5$ & 2.4h\\ 
            Full-FT & 10000  & - & - & 7.01$\times 10^5$ & 7.01$\times 10^5$  & 3.1h \\ 
\bottomrule
\end{tabular}
  }
	\label{tab:flops}
\end{table} 

\paragraph{Efficiency analysis}
We analyze the overhead of VLM in each data selection loop in Table \ref{tab:flops}. We estimate the backward computations triple the forward pass following the PyTorch report~\cite{li2020pytorch}. The CLIP-ViT-L model we use requires about 11$\times$ the FLOPs of DeiT, but only needs to be inferenced once in the loop. This leads to the FLOPs overhead of CEA to be merely 4.5\% of the finetuning cost. The resulting FLOPs reduction ratio is therefore similar to the batch number reduction ratio. Note that this analysis is consistent across datasets as all data are resized to 224$\times$224. 

\paragraph{Ablation study}
\label{subsection:ablation_study}
We first verify the importance of proposed techniques, the Objective-aware Data Selection (ODS) and Cross-attentive Embedding Augmentation (CEA) in Table \ref{ablation}. Specifically, ODS can be disabled by removing the $\mathcal{L}$ term in Equation \ref{equ:objective}. ODS significantly improves model performance, especially with limited data, enhancing classification accuracy and expediting the finetuning process. CEA further improves classification accuracy by integrating rich semantic information from text embeddings, enhancing model generalization and capturing underlying semantics. 
We further study the impact of the number of loops using CIFAR-10 dataset in Table \ref{tab:loop}. The total number of training batches is the same in all experiments with equal division by the loop count. It is clear that the model performance is enhanced with the increase in the number of loops. However, performing additional loops of data selection leads to overhead in the total training time. We set number of loop to three throughout our experiments to balance the trade-off. Detailed analysis on the training time is provided in Appendix C.2.

\setlength\tabcolsep{3pt}
\begin{table}[t]
\footnotesize
	\centering
 	\caption{Ablation study of the number of data selection loops on the CIFAR-10 dataset with 5\% data selection.}
	\resizebox{0.95\columnwidth}{!}{
		\begin{tabular}{c|cccc}
\toprule
    \# Loops   & 2 & 3 & 4 & 5  \\
\midrule
            CIFAR & 95.89\tiny$\pm$0.02 & 96.01\tiny$\pm$0.02 & 96.14\tiny$\pm$0.03 & \textbf{96.24}\tiny$\pm$0.03 \\
\bottomrule
		\end{tabular}
  }
	\label{tab:loop}
 \vspace{-0.3cm}
\end{table} 

\section{Conclusions and future work}
In this paper, we improved the efficiency of finetuning a PVM towards a user-specified performance target with a novel active data selection framework, VeCAF. VeCAF finds a subset of training data that leads to faster convergence with an objective-aware data selection model, and additionally utilizes the text-domain knowledge of pretrained VLM to augment image embeddings. Through extensive experiments, we demonstrated the superior performance of the proposed approach when compared to baseline active data selection methods and the finetuning with all data.
In future work, we aim to further unleash the potential of text-domain augmentation by improving certain finegrained performance metrics in vision domain tasks, and by extending active data selection to active data generation for additional performance and efficiency improvements.

\clearpage

{\small
\bibliographystyle{ieee_fullname}
\bibliography{egbib}
}

\clearpage

\appendix
\noindent\textbf{Appendix}
\setlength\tabcolsep{3pt}
\begin{table*}[t]
	\centering
 	\caption{Convergence accuracy with unlimited number of batches. }
	\resizebox{\linewidth}{!}{%
		\begin{tabular}{c|c|ccc|ccc|ccc}
  \toprule
             \multirow{2}*{Method} &  \multirow{2}*{Loop}    & \multicolumn{3}{c|}{CIFAR-10} & \multicolumn{3}{c|}{Caltech101}  & \multicolumn{3}{c}{ImageNet-1K}\\
             \multirow{2}{*}{} & \multirow{2}{*}{} &
            1\% & 2\% & 5\% & 2\% & 5\% & 10\% & 1\% & 2\% & 4\%   \\
\midrule
            Full Data FT & single-run & \multicolumn{3}{c|}{99.31\tiny$\pm$0.01} & \multicolumn{3}{c|}{88.24\tiny$\pm$0.02} & \multicolumn{3}{c}{82.76\tiny$\pm$0.01} \\
\midrule
            \multirow{2}*{LearnLoss\cite{yoo2019learning}} & single-run & 90.07\tiny$\pm$0.02 & 93.67\tiny$\pm$0.03 & 95.99\tiny$\pm$0.02 & 62.88\tiny$\pm$0.02 & 73.09\tiny$\pm$0.03 & 83.04\tiny$\pm$0.04 & 52.97\tiny$\pm$0.03 & 60.14\tiny$\pm$0.03 & 61.93\tiny$\pm$0.03\\
            \multirow{2}{*}{} & multi-run & 90.25\tiny$\pm$0.03 & 94.21\tiny$\pm$0.03 & 96.32\tiny$\pm$0.04 & 63.74\tiny$\pm$0.02 & 73.25\tiny$\pm$0.03 & 83.31\tiny$\pm$0.03 & 53.66\tiny$\pm$0.05 & 60.49\tiny$\pm$0.03 & 62.32\tiny$\pm$0.04\\
\midrule 
            \multirow{2}*{ActiveFT~\cite{xie2023active}} & single-run & 92.31\tiny$\pm$0.02 & 95.46\tiny$\pm$0.02 & 98.18\tiny$\pm$0.04 & 73.69\tiny$\pm$0.02 & 81.33\tiny$\pm$0.03 & 86.78\tiny$\pm$0.02 & 56.87\tiny$\pm$0.04 & 63.19\tiny$\pm$0.03 & 66.01\tiny$\pm$0.03\\
            \multirow{2}{*}{} & multi-run & 92.95\tiny$\pm$0.01 & 95.87\tiny$\pm$0.03 & 98.54\tiny$\pm$0.02 & 74.22\tiny$\pm$0.02 & 81.88\tiny$\pm$0.03 & 87.04\tiny$\pm$0.02 & 57.11\tiny$\pm$0.03 & 63.46\tiny$\pm$0.03 & 66.21\tiny$\pm$0.02\\
\midrule
            VeCAF(ours) & multi-run & \textbf{93.87}\tiny$\pm$0.02 & \textbf{96.47}\tiny$\pm$0.01 & \textbf{98.97}\tiny$\pm$0.01 & \textbf{75.36}\tiny$\pm$0.01 & \textbf{83.62}\tiny$\pm$0.02 & \textbf{87.72}\tiny$\pm$0.01 & \textbf{59.41}\tiny$\pm$0.02 & \textbf{65.64}\tiny$\pm$0.01 & \textbf{68.52}\tiny$\pm$0.02\\
\bottomrule
		\end{tabular}
	}
	\label{main_result}
\end{table*}

In the supplementary material, we provide additional information for the main paper. We start by providing the datasets information in ~\cref{appendix:datasets} and additional details on our reproduction of previous active learning baselines in~\cref{appendix:transplantation}, and explains the missing results indicated by ``-'' in the experiment section of the paper.
Additional insights into loss convergence on the CIFAR dataset are meticulously documented in ~\cref{appendix:loss}. We delve into the efficiency of our approach in ~\cref{appendix:time_consumption}, providing a compelling narrative on the method's expediency. Lastly, an extensive evaluation of the model's accuracy, achieved with an unrestricted count of training batches, is detailed in~\cref{appendix:convergence_accuracy}, solidifying the robustness of our experimental findings.

\setlength\tabcolsep{4pt}
\begin{table}[t]
\footnotesize
	\centering
 	\caption{Running time to
select various percentages of samples from the Caltech101 training set for each data selection loop.}
	\resizebox{\columnwidth}{!}{
		\begin{tabular}{c|cccc}
\toprule
           Sel. ratio  & ALFA-Mix & LearnLoss & ActiveFT & VeCAF  \\
\midrule
            2\%  & 6m45s  & 1m42s  & 12.02s & 16.38s\\
            10\% & 52m31s & 23m17s & 13.36s & 18.87s\\
\bottomrule
		\end{tabular}
  }
	\label{time}
\end{table} 

Besides this document, we also include the source code of VeCAF in the supplementary material. Please check the README file for details. 


\section{Datasets}
\label{appendix:datasets}
CIFAR-10~\cite{krizhevsky2009learning} consists of 60,000 images with a resolution of 32$\times$32 pixels, divided into 10 categories. The training set contains 50,000 images while the test set contains 10,000 images. The Caltech101~\cite{fei2004learning} dataset consists of images from 101 object categories with 40 to 800 images per category. Most classes contain around 50 images, and the image resolution is approximately 300$\times$200 pixels.
ImageNet-1K~\cite{deng2009imagenet} is a larger dataset with 1,331,167 images belonging to 1,000 classes. The training set consists of 1,281,167 images while the validation set contains 50,000 images. All the training sets from these datasets are considered candidate pools for selection. We also leverage the ImageNet-C~\cite{hendrycks2019benchmarking} as an OOD test set to evaluate our VeCAF under out-of-distribution scenarios. ImageNet-C is an openly available dataset that consists of algorithmically generated corruptions, including blur and noise, applied to the ImageNet test set. It serves as a valuable resource for evaluating the robustness and generalization capabilities of computer vision models.

\begin{figure}[t]
\includegraphics[width=\linewidth]{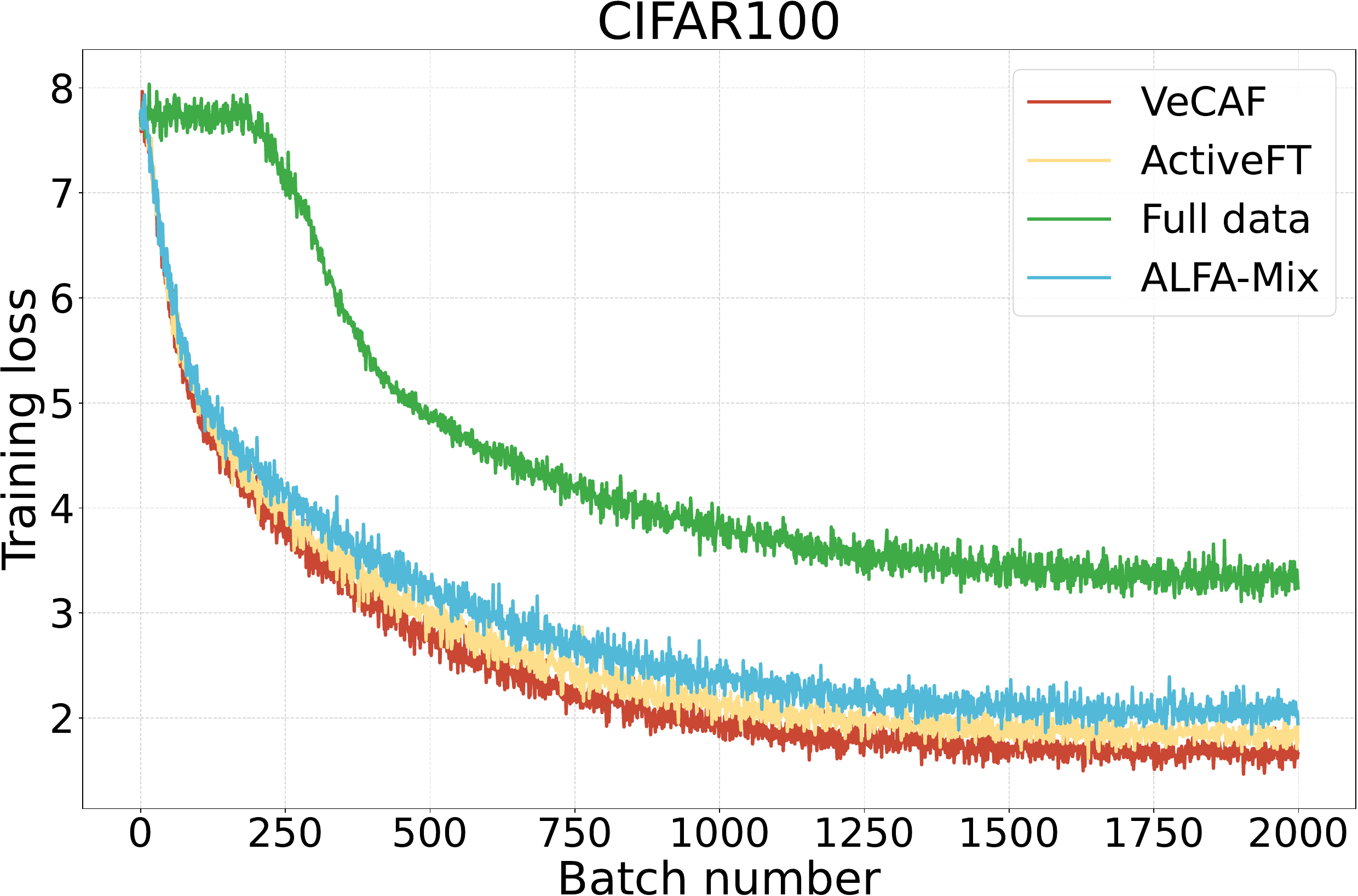}
\centering
\caption{Training curve of VeCAF and baselines including ActiveFT, ALFA-Mix, and Full data FT on 5\% CIFAR-100.}
\label{fig:loss}
\end{figure}

\section{Active learning reproducing details}
\label{appendix:transplantation}
In our study, we incorporate three well-established active learning algorithms, namely LearnLoss~\cite{yoo2019learning}, TA-VAAL~\cite{kim2021task}, and ALFA-Mix~\cite{parvaneh2022active}, within the pretraining-finetuning paradigm for image classification tasks. Specifically, we evaluate these algorithms on three widely used datasets, namely CIFAR-10, Caltech101, and ImageNet. To ensure a systematic and consistent evaluation, all three algorithms employ a batch-selection strategy for sample acquisition during the active learning process.

In Table 1 of our paper, the presence of ``-'' is attributed to the nature of traditional active learning methods, which require a small initial set randomly sampled at the beginning of the process. It is important to note that the performance of these active learning algorithms on this initial set is comparable to random sampling. Therefore, to avoid redundancy, we have omitted reporting duplicate results for these random initial sets. For instance, in the case of CIFAR-10, we exclude reporting results for the sampling ratio of 0.5\%, and for Caltech101, results for the sampling ratio of 1\% are not reported.
Moreover, for the ImageNet-1K dataset, the reporting results of sampling ratios of 1\% and 2\% are omitted as the size of the initial set is 2.5\% according to the reported setting of these previous papers. 
By excluding duplicate results for random initial sets and smaller sampling ratios, we aim to present clear and concise information in Table 1, focusing on the most relevant and informative performance metrics for the active learning methods applied in our study.

\section{Additional experiment results}
\label{appendix:experiment}
\subsection{Loss convergence analysis}
\label{appendix:loss}
Given that the loss convergence on CIFAR-10 data is sufficiently rapid for each baseline and thus may not effectively highlight the advantages of our proposed VeCAF, we instead present the loss convergence of CIFAR-100, which possesses analogous domain characteristics to CIFAR-10, as illustrated in Figure \ref{fig:loss}. Mirroring the depiction in Figure 4 of the primary text, Figure \ref{fig:loss} exhibits a comparable trend in loss convergence, where our proposed VeCAF not only converges more swiftly but also to a lower loss value in comparison to the baselines. This demonstration underscores the enhanced convergence efficiency of VeCAF, both in terms of speed and performance.

\subsection{Time complexity of data selection}
\label{appendix:time_consumption}
Efficiency is a crucial aspect of the VeCAF, and it is desirable for it to operate in a time-efficient manner, so as to reduce the overhead of data selection in each training loop. In our study, we evaluate the time required to select various proportions of training samples from the Caltech101 dataset with selection ratio 2\% and 10\%, as shown in~\ref{time}. Here we consider the image caption of each training sample readily available as they can be generated offline for only once, while the time for performing ODS, text embedding generation, and CEA are included in the reported VeCAF time. 
Traditional active learning algorithms like LearnLoss~\cite{yoo2019learning} and ALFA-Mix~\cite{parvaneh2022active} requires multiple trial model updates to gradually adjust the selected data, where these trial updates constitute the majority of the time in the data selection process, making them significantly inefficient. In contrast, ActiveFT and our proposed VeCAF method perform sample selection in a single pass at the beginning of each data selection loop, eliminating the need for performing trial updates on the model. This results in significant time savings compared to traditional approaches. The slight time increase of VeCAF over ActiveFT is caused by the text embedding CEA process. Meanwhile, considering the $>150$s model training time in each loop, this 4-5s (3\%) time overhead in negligible, and also justifiable considering the benefits offered by VeCAF.

\subsection{Accuracy under unlimited training batches}
\label{appendix:convergence_accuracy}
This work concentrates on an efficient training paradigm, and accordingly, we have presented most of our experimental outcomes in the primary manuscript using a limited number of batches. This approach inherently benefits methodologies that enable quicker convergence. To thoroughly assess VeCAF's convergence efficacy, we have lifted the constraints on the number of training batches in this section and conducted a comparative analysis of VeCAF's final convergence metrics against established active learning frameworks. The results, delineated in~\ref{main_result}, demonstrate that VeCAF not only achieves expedited convergence but also surpasses previous active learning strategies in terms of final performance metrics. Remarkably, VeCAF attains performance on par with comprehensive finetuning by utilizing merely 5\% of the data for CIFAR-10 and 10\% for Caltech101. These findings suggest that VeCAF is capable of significantly enhancing both computational and data efficiency throughout the PVM finetuning procedure.

\end{document}